\documentclass[runningheads]{llncs}
\usepackage{subcaption}
\usepackage[utf8]{inputenc}
\usepackage[english]{babel}
\usepackage{url}
\usepackage{amssymb}
\usepackage{amsmath}
\usepackage{array}
\usepackage{graphicx}
\usepackage{algorithmic}
\usepackage[misc]{ifsym}

\newcolumntype{P}[1]{>{\centering\arraybackslash}p{#1}}

\usepackage[export]{adjustbox}
\usepackage{verbatim}

\usepackage[norelsize,linesnumbered,ruled,vlined]{algorithm2e}

\begin{document}
\title{Scalable Nonlinear AUC Maximization Methods}
\toctitle{Scalable Nonlinear AUC Maximization Methods}
%
%
\author{\Letter{ Majdi Khalid} \and Indrakshi Ray \and Hamidreza Chitsaz}
\tocauthor{\Letter{ Majdi Khalid}, Indrakshi Ray, and Hamidreza Chitsaz}
%
%
\institute{Computer Science Department \\
Colorado State University, Fort Collins, USA \\
\email{majdi.khaled@gmail.com},\email{indrakshi.ray@colostate.edu}\\
\email{chitsaz@chitsazlab.org}}

\maketitle              
\begin{abstract}
The area under the ROC curve (AUC) is a widely used measure for evaluating classification performance on heavily imbalanced data. The kernelized AUC maximization machines have established a superior generalization ability compared to linear AUC machines because of their capability in modeling the complex nonlinear structures underlying most real-world data. However, the high training complexity renders the kernelized AUC machines infeasible for large-scale data. In this paper, we present two nonlinear AUC maximization algorithms that optimize linear classifiers over a finite-dimensional feature space constructed via the k-means Nystr\"{o}m approximation. Our first algorithm maximizes the AUC metric by optimizing a pairwise squared hinge loss function using the truncated Newton method. However, the second-order batch AUC maximization method becomes expensive to optimize for extremely massive datasets. This motivates us to develop a first-order stochastic AUC maximization algorithm that incorporates a scheduled regularization update and scheduled averaging to accelerate the convergence of the classifier. Experiments on several benchmark datasets demonstrate that the proposed AUC classifiers are more efficient than kernelized AUC machines while they are able to surpass or at least match the AUC performance of the kernelized AUC machines. We also show experimentally that the proposed stochastic AUC classifier is able to reach the optimal solution, while the other state-of-the-art online and stochastic AUC maximization methods are prone to suboptimal convergence.

\end{abstract}

\section{Introduction}

The area under the ROC Curve (AUC) \cite{hanley1982meaning} has a wide range of applications in machine learning and data mining such as recommender systems, information retrieval, bioinformatics, and anomaly detection \cite{chaudhuri2016recommending,liu2009learning,rendle2009learning,agarwal2005generalization,root2015learning}. Unlike error rate, the AUC metric does not consider the class distribution when assessing the performance of classifiers. This property renders the AUC a reliable measure to evaluate classification performance on heavily imbalanced datasets \cite{cortes2004auc}, which are not uncommon in real-world applications.

The optimization of the AUC metric aims to learn a score function that scores a random positive instance higher than any negative instance. Therefore, the AUC metric is a threshold-independent measure. In fact, it evaluates a classifier over all possible thresholds, hence eliminating the effect of imbalanced class distribution. The objective function maximizing the AUC metric optimizes a sum of pairwise losses. This objective function can be solved by learning a binary classifier on pairs of positive and negative instances that constitute the difference space. Intuitively, the complexity of such algorithms increases linearly with respect to the number of pairs. However, linear ranking algorithms like RankSVM \cite{chapelle2010efficient,lee2014large}, which can optimize the AUC directly, have shown a learning complexity independent from the number of pairs.

However, the kernelized versions of RankSVM \cite{joachims2005support,chapelle2010efficient,kuo2014large} are superior to linear ranking machines in terms of producing higher AUC classification accuracy. This is due to its ability to model the complex nonlinear structures that underlie most real-world data. Analogous to kernel SVM, the kernelized RankSVM machines entail computing and storing a kernel matrix, which grows quadratically with the number of instances. This hinders the efficiency of kernelized RankSVM machines for learning on large datasets. 

The recent approaches attempt to scale up the learning for AUC maximization from different perspectives. The first approach adopts online learning techniques to optimize the AUC on large datasets \cite{kotlowski2011bipartite,zhao2011online,gao2013one,ding2015adaptive,khalid2016confidence}. However, online methods result in inferior classification accuracy compared to batch learning algorithms. The authors of \cite{kakkar2017sparse} develop a sparse batch nonlinear AUC maximization algorithm, which can scale to large datasets, to overcome the low generalization capability of online AUC maximization methods. However, sparse algorithms are prone to the under-fitting problem due to the sparsity of the model, especially for large datasets. The work in \cite{szorenyi2017non} imputes the low generalization capability of online AUC maximization methods to the optimization of the surrogate loss function on a limited hypothesis space. Therefore, it devises a nonparametric algorithm to maximize the real AUC loss function. However, learning such nonparametric algorithm on high dimensional space is not reliable.

In this paper, we address the inefficiency of learning nonlinear kernel machines for AUC maximization. We propose two learning algorithms that learn linear classifiers on a feature space constructed via the k-means Nystr\"{o}m approximation \cite{zhang2008improved}. The first algorithm employs a linear batch classifier \cite{chapelle2010efficient} that optimizes the AUC metric. The batch classifier is a Newton-based algorithm that requires the computation of all gradients and the Hessian-vector product in each iteration. While this learning algorithm is applicable for large datasets, it becomes expensive for training enormous datasets embedded in a large dimensional feature space. This motivates us to develop a first-order stochastic learning algorithm that incorporates the scheduled regularization update \cite{bordes2009sgd} and scheduled averaging \cite{polyak1992acceleration} to accelerate the convergence of the classifier. The integration of these acceleration techniques allows the proposed stochastic method to enjoy the low complexity of classical first-order stochastic gradient algorithms and the fast convergence rate of second-order batch methods.

The remainder of this paper is organized as follows. We begin by reviewing closely related work in Section 2. In Section 3, we define the AUC problem and present related background. The proposed methods are presented in Section 4. The experimental results are shown in Section 5. Finally, we conclude the paper and point out the future work in Section 6.

\section{Related Work}

The maximization of the AUC metric is a bipartite ranking problem, a special type of ranking algorithm. Hence, most ranking algorithms can be used to solve the AUC maximization problem. The large-scale kernel RankSVM is proposed in \cite{kuo2014large} to address the high complexity of learning kernel ranking machines. However, this method still depends quadratically on the number of instances, which hampers its efficiency. Linear RankSVM \cite{sculley2009large,chapelle2010efficient,lee2014large,airola2011training,joachims2006training} is more applicable to scaling up in comparison to the kernelized variations. However, linear methods are limited to linearly separable problems. Recent study \cite{chen2017ranking} explores the Nystr\"{o}m approximation to speed up the training of the nonlinear kernel ranking function. This work does not address the AUC maximization problem. It also does not consider the k-means Nystr\"{o}m method and only uses a batch ranking algorithm. Another method \cite{kakkar2017sparse} attempts to speed up the training of nonlinear AUC classifiers by learning a sparse model constructed incrementally based on chosen criteria \cite{keerthi2006building}. However, the sparsity can deteriorate the generalization ability of the classifier.

Another class of research proposes using online learning methods to reduce the training time required to optimize the AUC objective function \cite{kotlowski2011bipartite,zhao2011online,gao2013one,ding2015adaptive,khalid2016confidence}. The work in \cite{zhao2011online} addresses the complexity of pairwise learning by deploying a first-order online algorithm that maintains a buffer of fixed size for positive and negative instances. The work in \cite{khalid2016confidence} proposes a second-order online AUC maximization algorithm with a fixed-sized buffer. The work \cite{gao2013one} maintains the first-order and second-order statistics for each instance instead of the buffering mechanism. Recently the work in \cite{ying2016stochastic} formulates the AUC maximization problem as a convex-concave saddle point problem. The proposed algorithm in \cite{ying2016stochastic} solves a pairwise squared hinge loss function without the need to access the buffered instances or the second-order information. Therefore, it shows linear space and time complexities per iteration with respect to the number of features.

The work in \cite{hu2015kernelized} proposes a budget online kernel method for nonlinear AUC maximization. For massive datasets, however, the size of the budget needs to be large to reduce the variance of the model and to achieve an acceptable accuracy, which in turns increases the training time complexity. The work \cite{ding2017large} attempts to address the scalability problem of kernelized online AUC maximization by learning a mini-batch linear classifier on an embedded feature space. The authors explore both Nystr\"{o}m approximation and random Fourier features to construct an embedding in an online setting. Despite their superior efficiency, online linear and nonlinear AUC maximization algorithms are susceptible to suboptimal convergence, which leads to inferior AUC classification accuracy.

Instead of maximizing a surrogate loss function, the authors of \cite{szorenyi2017non} attempt to optimize the real AUC loss function using a nonparametric learning algorithm. However, learning the nonparametric algorithm on high dimensional datasets is not reliable.
 
\section{Preliminaries and Background}

\subsection{Problem Setting}
Given a training dataset $\mathcal{S} = \{x_{i},y_{i}\} \in \mathbb{R}^{n\times d}$, where $n$ denotes the number of instances and $d$ refers to the dimension of the data, generated from unknown distribution $\mathcal{D}$. The label of the data is a binary class label $y=\{-1,1\}$. We use $n^{+}$ and $n^{-}$ to denote the number of positive and negative instances, respectively. The maximization of the AUC metric is equivalent to the minimization of the following loss function:

\begin{eqnarray}\label{eq1}
\mathcal{L}(f;\mathcal{S}) = \frac{1}{n} \sum_{i=1}^{n^{+}} \sum_{j=1}^{n^{-}}  I(f(x^{+}_{i}) \leq f(x^{-}_{j})),
\end{eqnarray}

\noindent
for a linear classifier $f(x) = w^{T}x$, where $I(\cdot)$ is an indicator function that outputs $1$ if its argument is true, and $0$ otherwise. The discontinuous nature of the indicator function makes the pairwise minimization problem (\ref{eq1}) hard to optimize. It is common to replace the indicator function with its convex surrogate function as follows,

\begin{eqnarray}\label{eq2}
\mathcal{L}(f;\mathcal{S}) = \frac{1}{n} \sum_{i=1}^{n^{+}} \sum_{j=1}^{n^{-}} \ell(f(x^{+}_{i}) - f(x^{-}_{j}))^{p}.
\end{eqnarray}

This pairwise loss function $\ell(f(x^{+}_{i}) - f(x^{-}_{j}))$ is convex in $w$, and it upper bounds the indicator function. The pairwise loss function is defined as hinge loss when $p=1$, and is defined as squared hinge loss when $p=2$. The optimal linear classifier $w$ for maximizing the AUC metric can be obtained by minimizing the following objective function:

\begin{eqnarray}\label{eq3}
 \underset{w}{\operatorname{min}}  \frac{1}{2} ||w||^{2} + C \sum_{i=1}^{n^{+}} \sum_{j=1}^{n^{-}} max(0,1 - w^{T}(x^{+}_{i} - x^{-}_{j}))^{p},
\end{eqnarray}

\noindent
where $||w||$ is the Euclidean norm and $C$ is the regularization hyper-parameter. Notice that the weight vector $w$ is trained on the pairs of instances $(x^{+} - x^{-})$ that form the difference space. This linear classifier is efficient in dealing with large-scale applications, but its modeling capability is limited to the linear decision boundary.

The kernelized AUC maximization can also be formulated as an unconstrained objective function \cite{kuo2014large,chapelle2010efficient}:

\begin{equation}\label{eq4}
     \min_{\beta \in {\mathbb{R}^{n}}}  \frac{1}{2} \, \beta^{T} \,K\, \beta + C \sum_{(i,j)\in A} max(0,1 - ((K\beta)_{i}-(K\beta)_{j})^{p},
\end{equation}

\noindent
where $K$ is the kernel matrix, and $A$ is a sparse matrix that contains all possible pairs $A \equiv \{(i,j) | y_{i} > y_{j}\}$ . In the batch setting, the computation of the kernel costs $\mathcal{O}(n^{2}d)$ operations, while storing the kernel matrix requires $\mathcal{O}(n^{2})$ memory. Moreover, the summation over pairs costs $\mathcal{O}(n \log n)$ \cite{kuo2014large}. These complexities make kernel machines costly to train compared to the linear model that has linear complexity with respect to the number of instances.

\subsection{Nystr\"{o}m Approximation}

The Nystr\"{o}m approximation \cite{kumar2009ensemble,zhang2008improved} is a popular approach to approximate the feature maps of linear and nonlinear kernels. Given a kernel function $K(\cdot,\cdot)$ and landmark points $\{u_{l}\}^{v}_{l=1}$ generated or randomly chosen from the input space $S$, the Nystr\"{o}m method approximates a kernel matrix $G$ as follows,

\begin{equation*}
  G \approx \bar{G} = E W^{-1} E^{T},
\end{equation*}

\noindent
where $W_{ij} = \kappa(u_{i},u_{j})$ is a kernel matrix computed on landmark points and $W^{-1}$ is its pseudo-inverse. The matrix $E_{ij} = \kappa(x_{i},u_{j})$ is a kernel matrix representing the intersection between the input space and the landmark points. The matrix $W$ is factorized using singular value decomposition or eigenvalue decomposition as follows: $W = U \Sigma^{-1} U^{T}$, where the columns of the matrix $ U $ hold the orthonormal eigenvectors while the diagonal matrix $ \Sigma $ holds the eigenvalues of $W$ in descending order. The Nystr\"{o}m approximation can be utilized to transform the kernel machines into linear machines by nonlinearly embedding the input space in a finite-dimensional feature space. The nonlinear embedding for an instance $x$ is defined as follows,  

\begin{equation*}
  \varphi(x) = U_{r} \; \Sigma_{r}^{-\frac{1}{2}} \phi^{T}(x),
\end{equation*}

\noindent
where $\phi(x) = [\kappa(x,u_{1}),\dots, \kappa(x,u_{v})]$, the diagonal matrix $\Sigma_{r}$ holds the top $r$ eigenvalues, and $U_{r}$ is the corresponding eigenvectors. The rank-$r$, $r \leq v$, is the best rank-$r$ approximation of $W$. We use the k-means algorithm to generate the landmark points \cite{zhang2008improved}. This method has shown a low approximation error compared to the standard method, which selects the landmark points based on uniform sampling without replacement from the input space. The complexity of the k-means algorithm is linear $\mathcal{O}(nvd)$, while the complexity of singular value decomposition or eigenvalue decomposition is $\mathcal{O}(v^{3})$. Therefore, the complexity of the k-means Nystr\"{o}m approximation is linear in the input space.

\section{Nonlinear AUC Maximization}
In this section, we present the two nonlinear algorithms that maximize the AUC metric over a finite-dimensional feature space constructed using the k-means Nystr\"{o}m approximation \cite{zhang2008improved}. First, we solve the pairwise squared hinge loss function in a batch learning mode using the truncated Newton solver \cite{chapelle2010efficient}. For the second method, we present a stochastic learning algorithm that minimizes the pairwise hinge loss function.

The main steps of the proposed nonlinear AUC maximization methods are shown in Algorithm \ref{alg1}. In the embedding steps, we construct the nonlinear mapping (embedding) based on a given kernel function and landmark points. The landmark points are computed by the k-means clustering algorithm applied to the input space. Once the landmark points are obtained, the matrix $W$ and its decomposition are computed. The original input space is then mapped nonlinearly to a finite-dimensional feature space in which the nonlinear problem can be solved using linear machines.

\begin{algorithm}[t]
\caption{Nonlinear AUC Maximization} \label{alg1}
\begin{algorithmic}
  
  \STATE {\bf Embedding Steps}: 
	\begin{itemize}
        Compute the centroid points $\{u_{l}\}^{v}_{l=1}$ \\
        Form the matrix  $W$: $W_{ij} = \kappa(u_{i},u_{j})$ \\
        Compute the eigenvalue decomposition: $W = U\Sigma U^{T}$ \\
        Form the matrix  $E$: $E_{i} = \phi(x_{i}) = [\kappa(x_{i},u_{1}),\dots ,\kappa(x_{i},u_{v})]$ \\
     Construct the feature space: $\varphi(X) = U_{r} \Sigma_{r}^{-\frac{1}{2}} E^{T}$ \\
	\end{itemize}
    
     \STATE {\bf Training}: 
     \begin{itemize}

     Learn the batch model described in Algorithm \ref{alg2} or the stochastic model detailed in Algorithm \ref{alg3}
     \end{itemize}
     
      \STATE {\bf Prediction}: 
      \begin{itemize}
      Map a test point $x$: $\varphi(x) = U_{r} \Sigma_{r}^{-\frac{1}{2}} \phi^{T}(x)$  \\
      Score value: $w^{T}\varphi(x)$ \\
      \end{itemize}
\end{algorithmic}
\end{algorithm}

The AUC optimization (\ref{eq3}) can be solved for $w$ in the embedded space as follows,

\begin{eqnarray}\label{eq5}
 \underset{w}{\operatorname{min}}  \frac{1}{2} ||w||^{2} + C \sum_{i=1}^{n^{+}} \sum_{j=1}^{n^{-}} max(0,1 - w^{T}(\varphi({x}^{+}_{i}) - \varphi(x^{-}_{j})))^{p},
\end{eqnarray}

\noindent
where $\varphi(x)$ is a nonlinear feature mapping for $x$. The minimization of (\ref{eq5}) can be solved using truncated Newton methods \cite{chapelle2010efficient} as shown in Algorithm \ref{alg2}. The matrix $A$ in Algorithm \ref{alg2} is a sparse matrix of size $r \times n$, where $r$ is the number of pairs. The matrix $A$ holds all possible pairs in which each row of $A$ has only two nonzero values. That is, if $(i,j) \;|\; y_{i} > y_{j}$, the matrix A has a $k$-th row such that $A_{ki}=1, A_{kj}=-1$. However, the complexity of this Newton batch learning is dependent on the number of pairs. The authors of \cite{chapelle2010efficient} also proposed the PSVM+ algorithm, which avoids the direct computation of pairs by reformulating the pairwise loss function in such a way that the calculations of the gradient and the Hessian-vector product are accelerated.

\begin{algorithm}[H]
\caption{Batch Nonlinear AUC Maximization} \label{alg2}

\begin{algorithmic}
\STATE \textbf{Input:} embedded data $\tilde{X}$
\STATE \textbf{Output:} the ranking model $w$

\STATE initial vector $w {\leftarrow} 0$
\WHILE {stopping criterion is not satisfied}
	\STATE $D = max(0, 1 - A (w^{T}\tilde{X}))$ 
	\STATE Compute gradient $g = w - (C D^{T} A \tilde{X})^{T}$ 
	\STATE Compute a search direction $s_{t}$ by applying conjugate gradient to solve $\nabla^{2}F(w_{k})s = - \nabla F(w_{k})$
	\STATE Update $w_{k+1} = w_{k} + s_{k} $
\ENDWHILE

\end{algorithmic}
\end{algorithm}

\bigskip
Nevertheless, the optimization of PRSVM+ to maximize the AUC metric still requires $O(n\hat{d}+2n+\hat{d})$ operations to compute each of the gradient and the Hessian-vector product in each iteration, where $\hat{d}$ is the dimension of the embedded space. This makes the training of PRSVM+ expensive for massive datasets embedded using a large number of landmark points. A large set of landmark points is desirable to improve the approximation of the feature maps; hence boosting the generalization ability of the involved classifier.

To address this complexity, we present a first-order stochastic method to maximize the AUC metric on the embedded space. Specifically, we optimize a pairwise hinge loss function using stochastic gradient descent accelerated by scheduling both the regularization update and averaging techniques. The proposed stochastic algorithm can be seen as an averaging variant of the SVMSGD2 method proposed in \cite{bordes2009sgd}. Algorithm \ref{alg3} describes the proposed stochastic AUC maximization method. The algorithm randomly selects a positive and negative instance and updates the model in each iteration as follows,

\begin{equation*}
 w_{t+1} = w_{t} + \frac{1}{\lambda (t+t_{0})} \ell^{\prime}(w^{T}_{t}x_{t})x_{t},
\end{equation*}

\noindent
where $\ell^{\prime}(z)$ is a subgradient of the hinge loss function, the vector $x_{t}$ holds the difference $\varphi({x}^{+}_{i}) - \varphi(x^{-}_{j})$, $w_{t}$ is the solution after $t$ iterations, and $\lambda (t+t_{0})$ is the learning rate, which decreases in each iteration. The hyper-parameter $\lambda$ can be tuned on a validation set. The positive constant $t_{0}$ is set experimentally, and it is utilized to prevent large steps in the first few iterations \cite{bordes2009sgd}. The model is regularized each $rskip$ iterations to accelerate its convergence. We also foster the acceleration of the model by implementing an averaging technique \cite{polyak1992acceleration,xu2011towards}. The intuitive idea behind the averaging step is to reduce the variance of the model that stems from its stochastic nature. We regulate the regularization update and averaging steps to be performed each $askip$ and $rskip$ iterations as follows,

\begin{equation*}
w_{t+1} = w_{t+1} - rskip(t+t_{0})^{-1} w_{t+1}
\end{equation*}

\begin{equation*}
\tilde{w}_{q+1} = \frac{q \tilde{w}_{q} +  w_{t+1}}{q + 1},
\end{equation*}

\noindent
where $\tilde{w}$ is the averaged solution after $q$ iterations with respect to the $askip$. The advantage of regulating the averaging step is to reduce the per iteration complexity, while effectively accelerating the convergence. 

The presented first-order stochastic AUC maximization requires $\mathcal{O}(\hat{d}a)$ operations per iteration in addition to the $\mathcal{O}(\hat{d})$ operations needed for each of the regularization update and averaging steps that occur per $rskip$ and $askip$ iterations respectively, where $a$ denotes the average number of nonzero coordinates in the embedded difference vector $x_{t}$.

\begin{algorithm}[H]
\caption{Stochastic Nonlinear AUC Maximization} \label{alg3}

\begin{algorithmic}
\STATE \textbf{Input:} embedded data $\tilde{X}$, $\lambda$, $t_{0}$,$T$, rskip, askip 
\STATE \textbf{Output:} the ranking model $w$
 \STATE $w_{1} {\leftarrow} 0$ and $\tilde{w}_{0} {\leftarrow} 0,$$rcount = rskip$, $acount = askip$, $q = 0$
\FOR {$t = 1, \dots, T$}
  
  \STATE Randomly pick a pair $i_{t} \in 1,\dots, n^{+},j_{t} \in 1,\dots, n^{-}$ 
  
  \STATE $x_{t} = \tilde{x}_{i_{t}} - \tilde{x}_{j_{t}}$
  
  \STATE $w_{t+1} = w_{t} + \frac{1}{\lambda (t+t_{0})} \ell^{\prime}(w^{T}_{t}x_{t})x_{t}$ 
  
  \STATE $rcount = rcount - 1$ 
  
  \IF{rcount $\leq$ 0}
  
  	\STATE $w_{t+1} = w_{t+1} - rskip(t+t_{0})^{-1} w_{t+1}$
  
  	\STATE $rcount = rskip$
   
  \ENDIF
  
  \STATE $acount = acount - 1$ 
  
  \IF{acount $\leq$ 0}
  	
  	\STATE $\tilde{w}_{q+1} = \frac{q \tilde{w}_{q} +  w_{t+1}}{q\;+\;1}$
    
    \STATE $q = q + 1$
  
  	\STATE acount = askip
   
  \ENDIF
\ENDFOR

\STATE set $w = \tilde{w}_{q}$
\STATE \textbf{return $w$}

\end{algorithmic}
\end{algorithm}

\section{Experiments}
In this section, we evaluate the proposed methods on several benchmark datasets and compare them with kernelized AUC algorithm and other state-of-the-art online AUC maximization algorithms. The experiments are implemented in MATLAB, while the learning algorithms are written in C language via MEX files. The experiments were performed on a computer equipped with an Intel 4GHz processor with 32G RAM.

\subsection{Benchmark Datasets}
The datasets we use in our experiments can be downloaded from LibSVM website\footnote{\url{https://www.csie.ntu.edu.tw/~cjlin/libsvmtools/datasets/}} or UCI\footnote{\url{http://archive.ics.uci.edu/ml/index.php}}. The datasets that are not split (i.e., spambase, magic04, connect-4, skin, and covtype) into training and test sets; we randomly divide them into 80$\%$-20$\%$ for training and testing. The features of each dataset are standardized to have zero mean and unit variance. The multi-class datasets (e.g., covtype and usps) are converted into class-imbalanced binary data by grouping the instances into two sets, where each set has the same number of class labels. To speed up the experiments that include the kernelized AUC algorithm, we train all the compared methods on 80k instances, randomly selected from the training set. The other experiments are performed on the entire training data. The characteristics of the datasets along with their imbalance ratios are shown in Table \ref{table1}.

\begin{table*}[h]
\caption{Benchmark datasets}
\label{table1}
\centering

\begin{tabular}{c|c c c c} \hline 

Data      &   \#training  &   \#test    &  \#feat      &  ratio    \\
 \hline

spambase & 3,680 & 921 & 57 & 1.53 \\

usps & 7,291 & 2,007 & 256 & 1.40 \\

magic04 & 15,216 & 3,804 & 10 &  1.84  \\

protein & 17,766 & 6,621 & 357 & 2.11 \\

ijcnn1 & 49,990 & 91,701 & 22 & 9.44 \\

connect-4 & 54,045 & 13,512 & 126 & 3.06 \\

acoustic & 78,823 & 19,705 & 50 & 3.31 \\

skin & 196,045 & 49,012 & 3 &  3.83 \\

cod-rna & 331,152 & 157,413 & 8 & 2.0  \\

covtype & 464,809 & 116,203 & 54 & 10.65 \\

\hline

\end{tabular}

\end{table*}

\subsection{Compared Methods and Model Selection}
We compare the proposed methods with kernel RankSVM and linear RankSVM, which can be used to solve the AUC maximization problem. We also include two state-of-the-art online AUC maximization algorithms. The random Fourier method that approximates the kernel function is also involved in the experiments where the resulting classifier is solved by linear RankSVM.   

\begin{enumerate}

\item \textbf{RBF-RankSVM:} 
This is the nonlinear kernel RankSVM \cite{kuo2014large}. We use Gaussian kernel $K(x,y) = exp(-\gamma ||x-y||^2)$ to model the nonlinearity of the data. The best width of the kernel $\gamma$ is chosen by 3-fold cross validation on the training set via searching in $\{2^{-6},\dots,2^{-1}\}$. The regularization hyper-parameter $C$ is also tuned by 3-fold cross validation by searching in the grid $\{2^{-5},\dots, 2^{5}\}$. The searching grids are selected based on \cite{kuo2014large}. We also train the RBF-RankSVM on $1/5$ subsamples, selected randomly. 

\item \textbf{Linear RankSVM (PRSVM+):} 
This is the linear RankSVM that optimizes the squared hinge loss function using truncated Newton \cite{chapelle2010efficient}. The best regularization hyper-parameter $C$ is chosen from the grid $\{2^{-15},\dots,2^{10}\}$ via 3-fold cross validation.

\item \textbf{RFAUC:} 
This uses the random Fourier features \cite{rahimi2008random} to approximate the kernel function. We use PRSVM+ to solve the AUC maximization problem on the projected space. The hyper-parameters C and $\gamma$ are selected via 3-fold cross validation by searching on the grids $\{2^{-15},\dots,2^{10}\}$ and $\{1,10,100\}$, respectively.

\item \textbf{NOAM:} 
This is the sequential variant of online AUC maximization \cite{zhao2011online} trained on a feature space constructed via the k-means Nystr\"{o}m approximation. The hyper-parameters are chosen as suggested by \cite{zhao2011online} via 3-fold cross validation. The number of positive and negative buffers is set to 100.

\item \textbf{NSOLAM:} 
This is the stochastic online AUC maximization \cite{ying2016stochastic} trained on a feature space constructed via the k-means Nystr\"{o}m approximation. The hyper-parameters of the algorithm (i.e., the learning rate and the bound on the weight vector) are selected via 3-fold cross validation by searching in the grids $\{1:9:100\}$ and $\{10^{-1},\dots,10^{5}\}$, respectively. The number of epochs is set to 15.

\item \textbf{NBAUC:} 
This is the proposed batch AUC maximization algorithm trained on the embedded space. We solve it using the PRSVM+ algorithm \cite{chapelle2010efficient}. The hyper-parameter $C$ is tuned similarly to the Primal RankSVM.

\item \textbf{NSAUC:} 
This is the proposed stochastic AUC maximization algorithm trained on the embedded space. The hyper-parameter $\lambda$ is chosen from the grid $\{10^{-10},\dots,10^{-7}\}$ via 3-fold cross validation.

\end{enumerate}

For those algorithms that involve the k-means Nystr\"{o}m approximation (i.e., our proposed methods, NOAM, and NSOLAM), we compute $1600$ landmark points using the k-means clustering algorithm, which is implemented in C language. We select a Gaussian kernel function to be used with the k-means Nystr\"{o}m approximation. The bandwidth of the Gaussian function is set to be the average squared distance between the first 80k instances and the mean computed over these 80k instances. For a fair comparison, we also set the number of random Fourier features to 1600.

\subsection{Results for Batch Methods}

The comparison of batch AUC maximization methods in terms of AUC classification accuracy on the test set is shown in Table \ref{table2}, while Table \ref{table3} compares these batch methods in terms of training time. For connect-4 dataset, the results of RBF-RankSVM are not reported because the training runs over five days.

We observe that the proposed NBAUC outperforms the competing batch methods in terms of AUC classification accuracy. The AUC performance of RBF-RankSVM might be improved for some datasets if the best hyper-parameters are selected on a more restricted grid of values. Nevertheless, the training of NBAUC is several orders of magnitude faster than RBF-RankSVM. The fast training of NBAUC is clearly demonstrated on the large datasets.

The proposed NBAUC shows a robust AUC performance compared to RFAUC on most datasets. This can be attributed to the robust capability of the k-means Nystr\"{o}m method in approximating complex nonlinear structures. It also indicates that a better generalization can be attained by capitalizing on the data to construct the feature maps, which is the main characteristic of the Nystr\"{o}m approximation, while the random Fourier features are oblivious to the data.

We also observe that the AUC performance of both RBF-RankSVM and its variant applied to random subsamples outperform the linear RankSVM, except for the protein dataset. However, RBF-RankSVM methods require longer training, especially for large datasets. We see that the linear RankSVM performs better than the kernel AUC machines on the protein dataset. This implies that the protein dataset is linearly separable. However, the AUC performance of the proposed method NBAUC is even better than linear RankSVM on this dataset.

\begin{table*}
\caption{Comparison of AUC performance for batch classifiers on the benchmark datasets.}
\label{table2}
\centering

\begin{tabular}{c|c|c|c|c|c} \hline

Data    &   RBF-RankSVM  & RBF-RankSVM\scriptsize{(subsample)} & Linear RankSVM  & RFAUC &\textbf{NBAUC} \\ \hline

spambase & 98.00  & 96.02 & 97.47  & 97.75 & 98.04  \\ 

usps & 99.08 & 98.54 & 90.27 &  97.42 & 99.24 \\

magic04 & 92.18 & 91.34 & 84.47 &   92.83 & 93.06  \\ 

protein & 80.97 & 77.60 & 83.30 &  58.43 & 84.33  \\ 

ijcnn1 & 99.68 & 99.35 & 91.56  & 98.86 & 99.57 \\ 

connect-4 & - & 91.32 & 88.20 & 91.10 & 94.09 \\

acoustic & 93.60 & 93.02 & 87.38 &  91.82 & 94.14 \\

skin & 99.92 & 99.92 & 94.81 &  100 & 99.98  \\ 

cod-rna & 99.07 & 99.07 & 98.85 &  99.12 & 99.12 \\

covtype & 93.94 & 94.05 & 87.75 &  95.99 & 96.03 \\ 

\hline

\end{tabular}

\end{table*}

 \begin{table*}
\caption{Comparison of training time (in seconds) for batch classifiers on the benchmark datasets.}
\label{table3}
\centering

\begin{tabular}{c|c|c|c|c|c} \hline

Data    &   RBF-RankSVM  & RBF-RankSVM\scriptsize{(subsample)} & Linear RankSVM  & RFAUC & \textbf{NBAUC} \\ \hline

spambase & 3.08  & 0.10 & 0.13 &  3.59 & 7.71  \\ 

usps & 492.30 & 0.83 & 1.42 &  6.77 & 27.68 \\

magic04 & 518.04 & 3.71 & 0.08 &  21.51 & 25.46  \\ 

protein & 2614.7 & 4.81 & 4.47 &  14.20 & 73.81  \\ 

ijcnn1 & 15,434 & 282 & 0.57  &  80.17 & 88.87 \\ 

connect-4 & - & 12,701 & 3.42 & 62.60 & 164.48 \\

acoustic & 134,030 & 5,610 & 1.88 & 92.74 & 151.78 \\

skin & 2037.30 & 78.20 & 0.20 & 73.18 & 23.71  \\ 

cod-rna & 5,715 & 255.4 & 0.44 &  83.01 & 113.66 \\

covtype & 133,270 & 11,670 & 2.54 & 273.67 & 220.90 \\ 

\hline

\end{tabular}

\end{table*}

\newpage

\subsection{Results for Stochastic Methods}

We now compare our stochastic algorithm NSAUC with the state-of-the-art online AUC maximization methods, NOAM and NSOLAM. We also include the results of the proposed batch algorithm NBAUC for reference. The k-means Nystr\"{o}m approximation is implemented separately for each algorithm as introduced in Section 4. We experiment on the following large datasets: ijcnn1, connect-4, acoustic, skin, cod-rna, and covtype. Table \ref{table4} shows the comparison of the proposed methods with the online AUC maximization algorithms. Notice that the reported training time in Table \ref{table4} indicates only the time cost of the learning steps with excluding the embedding steps. 

We can see that the proposed NSAUC achieves a competitive AUC performance compared to the proposed NBAUC, but with less training time. On the largest dataset covtype, the AUC performance of NSAUC is on par with NBAUC, while it only requires 49.17 seconds for training compared to more than 18 minutes required by NBAUC. In contrast to the online methods, the proposed NSAUC is able to converge to the optimal solution obtained by the batch method NBAUC. We attribute the robust performance of NSAUC to the effectiveness of scheduling both the regularization update and averaging. 

We observe that the proposed NSAUC requires longer training time on some datasets (e.g., connect-4 and acoustic) compared to the online methods; however, the difference in the training time is not significant. In addition, we see that NSOLAM performs better than NOAM in terms of AUC classification accuracy. This implies the advantage of optimizing the pairwise squared hinge loss function, performed by NSOLAM, over the pairwise hinge lose function, carried out by NOAM, for one-pass AUC maximization.

\begin{table*}[ht]
\centering

\caption{Comparison of AUC classification accuracy and training time (in seconds) for the proposed algorithms with other online AUC maximization algorithms. The training time does not include the embedding steps.}
\label{table4}
\begin{tabular}{l|l|l|l|l|l} \hline
Data      & Metric                                                            & NOAM                                                       & NSOLAM                                                     & \textbf{NSAUC}                                             & \textbf{NBAUC}                                             \\ \hline

ijcnn1    & \begin{tabular}[c]{@{}l@{}}AUC\\   \\ Training time\end{tabular} & \begin{tabular}[c]{@{}l@{}}98.16\\   \\ 6.24\end{tabular} & \begin{tabular}[c]{@{}l@{}}98.86\\   \\ 6.88\end{tabular} & \begin{tabular}[c]{@{}l@{}}99.69\\   \\ 
4.80\end{tabular} & \begin{tabular}[c]{@{}l@{}}99.57\\   \\ 40.70\end{tabular} \\ \hline

connect-4 & \begin{tabular}[c]{@{}l@{}}AUC\\ \\ Training time\end{tabular}    & \begin{tabular}[c]{@{}l@{}}85.96\\ \\ 6.97\end{tabular}   & \begin{tabular}[c]{@{}l@{}}90.60\\ \\ 7.39\end{tabular}   & \begin{tabular}[c]{@{}l@{}}94.04\\ \\ 10.74\end{tabular}   & \begin{tabular}[c]{@{}l@{}}94.08\\ \\ 36.96\end{tabular}  \\ \hline

acoustic  & \begin{tabular}[c]{@{}l@{}}AUC\\ \\ Training time\end{tabular}    & \begin{tabular}[c]{@{}l@{}}89.90\\ \\ 10.80\end{tabular}   & \begin{tabular}[c]{@{}l@{}}91.00\\ \\ 10.82\end{tabular}   & \begin{tabular}[c]{@{}l@{}}94.04\\ \\ 23.80\end{tabular}   & \begin{tabular}[c]{@{}l@{}}94.14\\ \\ 59.34\end{tabular}  \\ \hline

skin   & \begin{tabular}[c]{@{}l@{}}AUC\\ \\ Training time\end{tabular}    & \begin{tabular}[c]{@{}l@{}}99.98\\ \\ 6.26\end{tabular}   & \begin{tabular}[c]{@{}l@{}}99.01\\ \\ 5.66\end{tabular}   & \begin{tabular}[c]{@{}l@{}}99.98\\ \\ 6.60\end{tabular}   & \begin{tabular}[c]{@{}l@{}}99.98\\ \\ 10.32\end{tabular}   \\ \hline

cod-rna   & \begin{tabular}[c]{@{}l@{}}AUC\\    \\ Training time\end{tabular} & \begin{tabular}[c]{@{}l@{}}98.29\\ \\42.09\end{tabular}   & \begin{tabular}[c]{@{}l@{}}99.10\\ \\ 47.06\end{tabular}   & \begin{tabular}[c]{@{}l@{}}99.19\\ \\ 34.23\end{tabular}  & \begin{tabular}[c]{@{}l@{}}99.18\\ \\ 148.46\end{tabular}  \\ \hline

covtype   & \begin{tabular}[c]{@{}l@{}}AUC\\ \\ Training time\end{tabular}    & \begin{tabular}[c]{@{}l@{}}91.29\\ \\ 61.75\end{tabular}   & \begin{tabular}[c]{@{}l@{}}92.25\\ \\ 63.59\end{tabular}   & \begin{tabular}[c]{@{}l@{}}96.00\\ \\ 49.17\end{tabular}   & \begin{tabular}[c]{@{}l@{}}96.60\\ \\ 1110.44\end{tabular} 
\end{tabular}

\end{table*}

\newpage

\subsection{Study on the Convergence Rate} 
\begin{figure*}[ht]

\begin{subfigure}{0.3\textwidth}
\centering
\includegraphics[height=3cm,width=\linewidth]{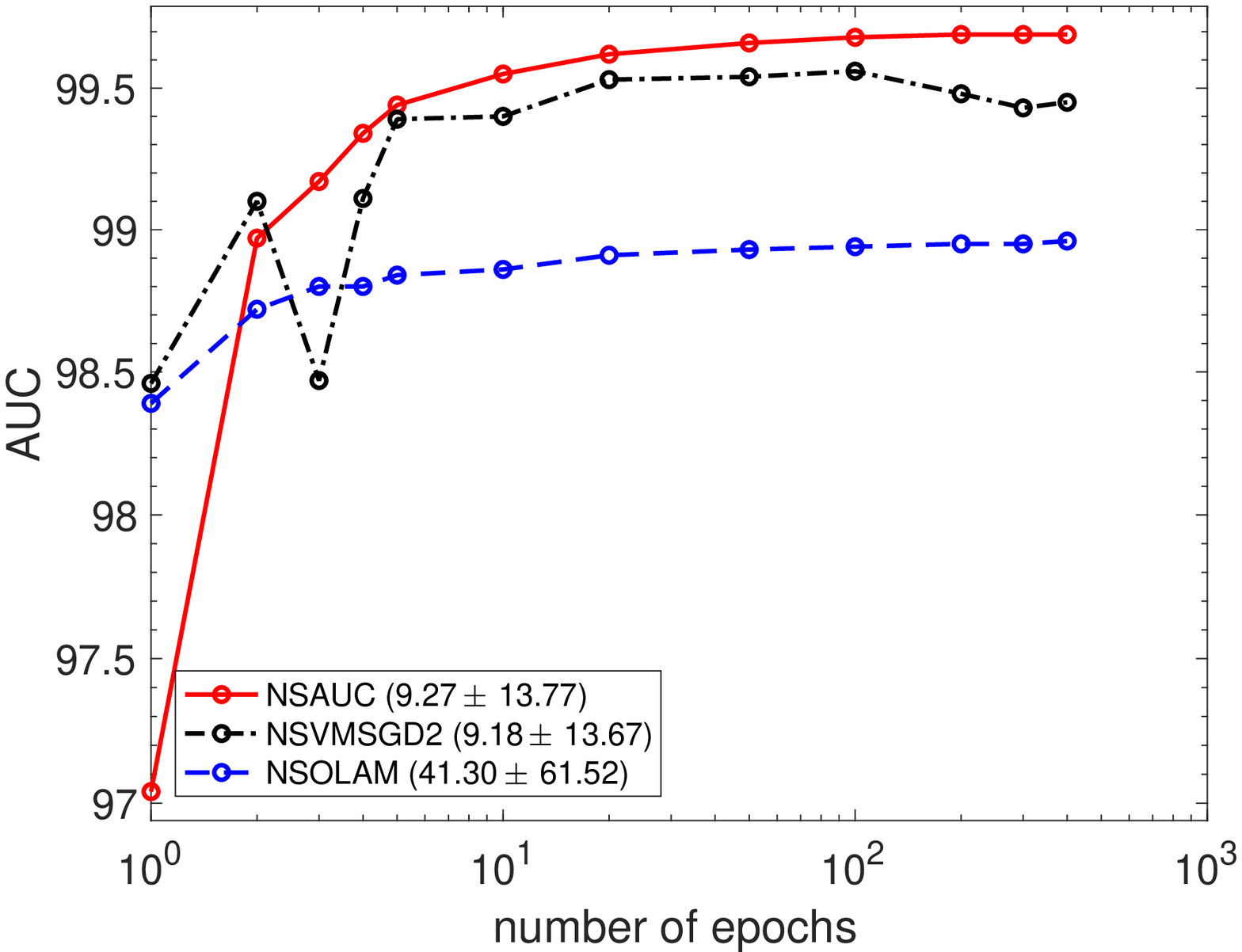}
\caption{ijcnn1}
\label{subfig:A}
\end{subfigure}\hfill
\begin{subfigure}{0.3\textwidth}
\centering
\includegraphics[height=3cm,width=\linewidth]{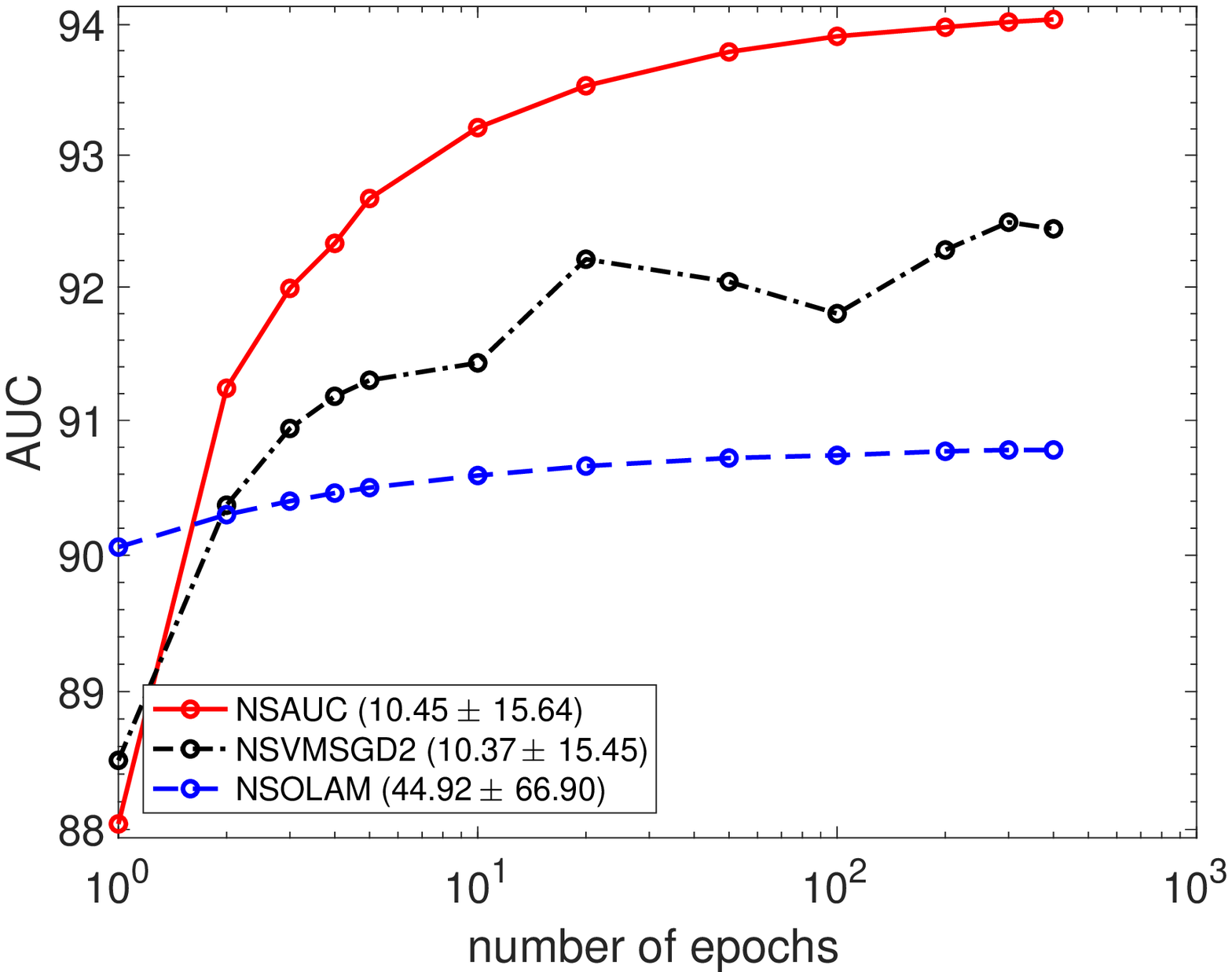}
\caption{connect-4}
\label{subfig:D}
\end{subfigure}\hfill
\begin{subfigure}{0.3\textwidth}
\centering
\includegraphics[height=3cm,width=\linewidth]{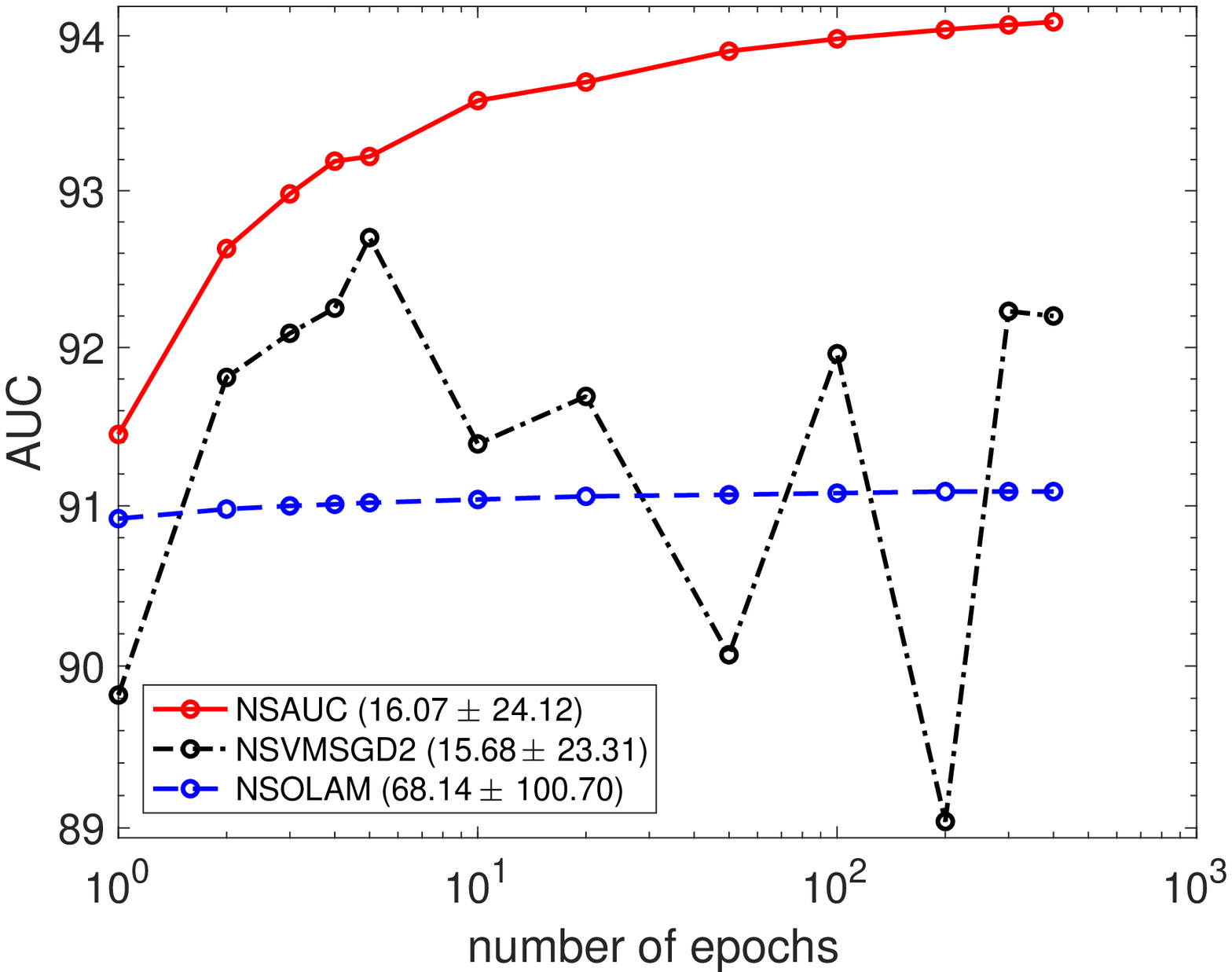}
\caption{acoustic}
\label{subfig:C}
\end{subfigure}\hfill
\begin{subfigure}{0.3\textwidth}
\centering
\includegraphics[height=3cm,width=\linewidth]{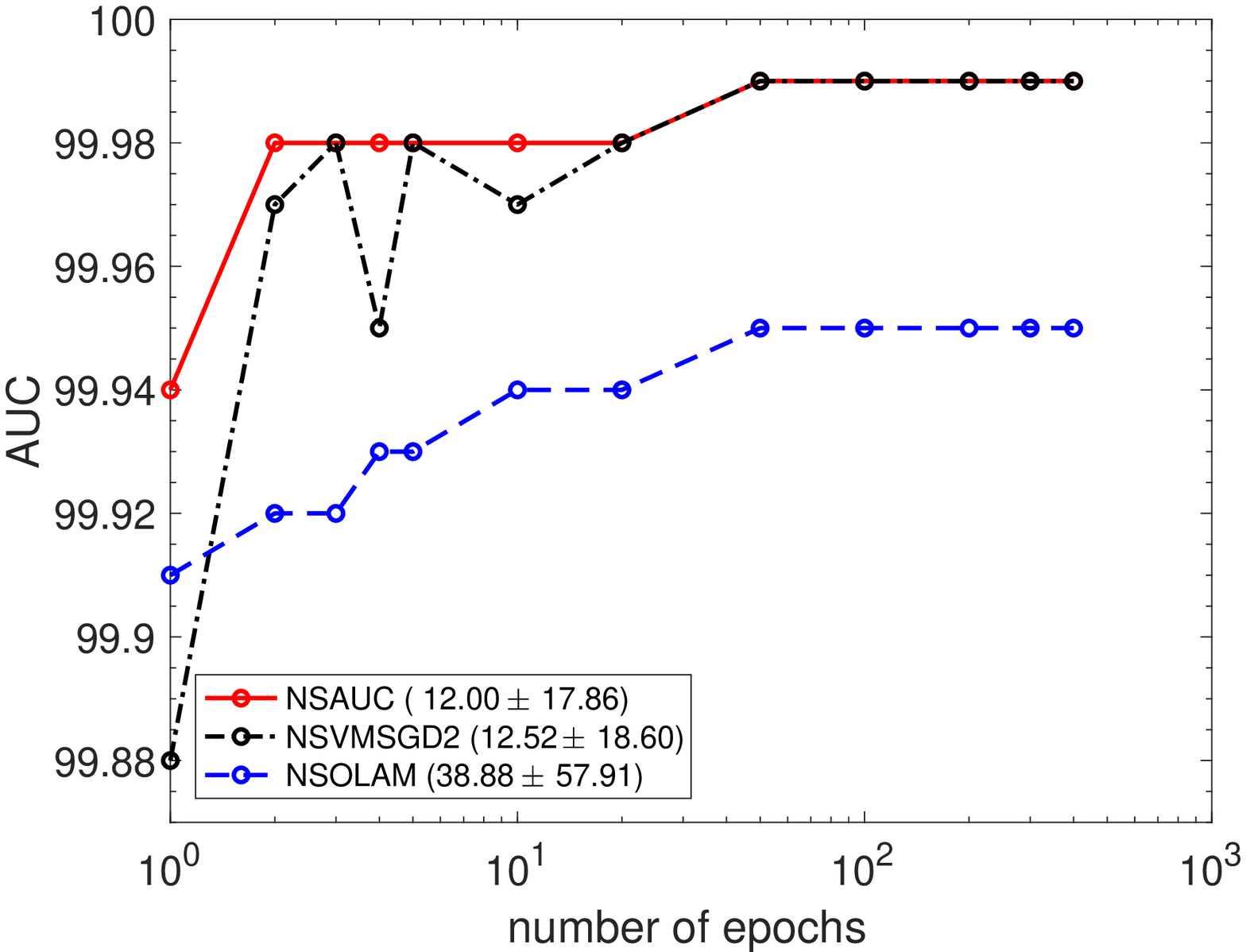}
\caption{skin}
\label{subfig:F}
\end{subfigure}\hfill
\begin{subfigure}{0.3\textwidth}
\centering
\includegraphics[height=3cm,width=\linewidth]{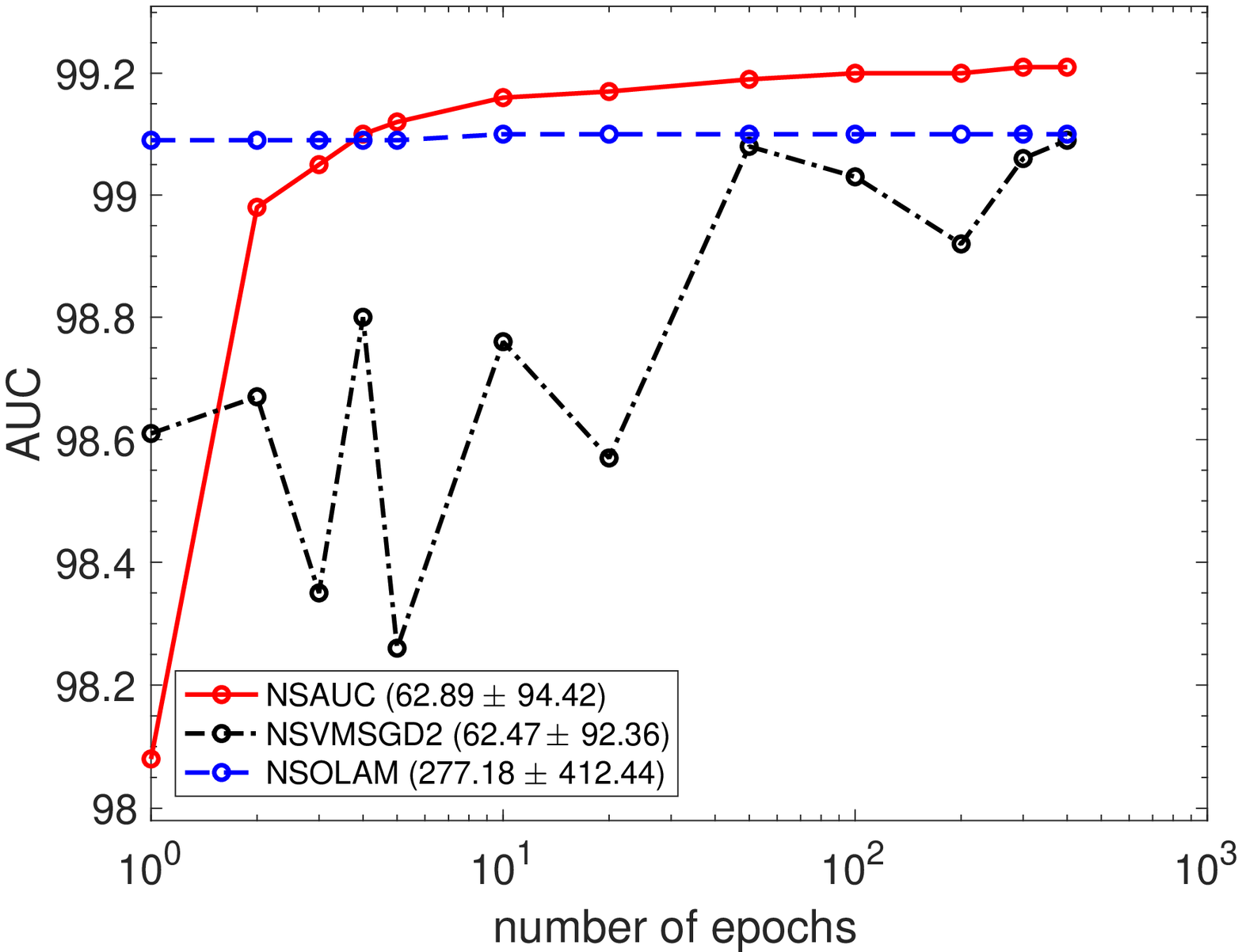}
\caption{cod-rna}
\label{subfig:B}
\end{subfigure}\hfill
\begin{subfigure}{0.3\textwidth}
\centering
\includegraphics[height=3cm,width=\linewidth]{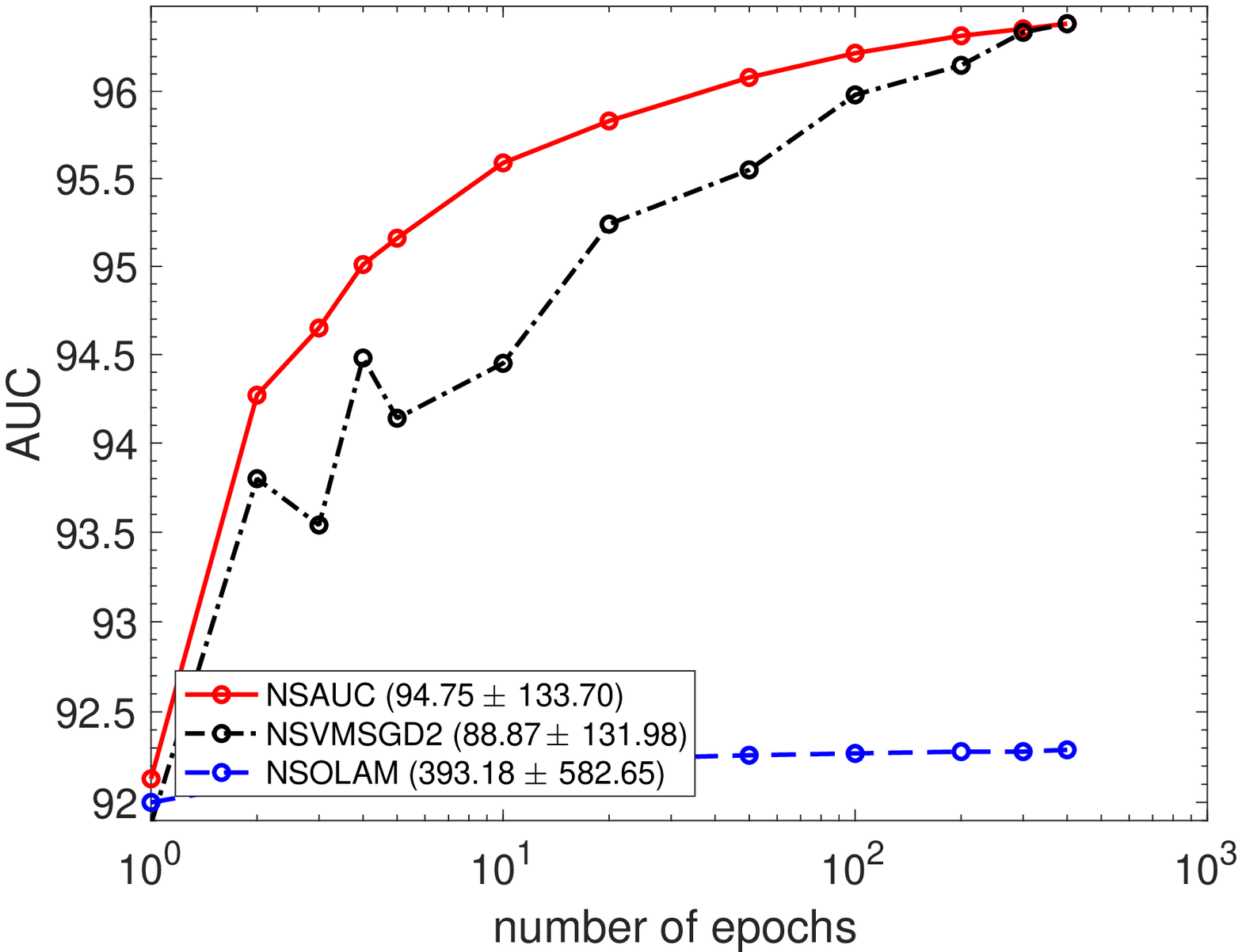}
\caption{covtype}
\label{subfig:E}
\end{subfigure}\hfill

 \caption{AUC classification accuracy of stochastic AUC algorithms with respect to the number of epochs. We randomly pick a positive and negative instance for each iteration in NSAUC and NSVMSGD2, where $n$ iterations correspond to one epoch. The values in parentheses denote the averaged training time (in seconds) along with the standard deviation over all epochs. The training time excludes the computational time of the embedding steps. The x-axis is displayed in log-scale.}
    \label{fig1}
  
\end{figure*}

We investigate the convergence of NSAUC and its counterpart NSOLAM with respect to the number of epochs. We also include NSVMSGD2 algorithm \cite{bordes2009sgd} that minimizes the pairwise hinge loss function on a feature space constructed via the k-means Nystr\"{o}m approximation, described in Section 4. The algorithm NSVMSGD2 is analogous to the proposed algorithm NSAUC, but with no averaging step. The AUC performances of these stochastic methods upon varying the number of epochs are depicted in Figure \ref{fig1}. We vary the number of epochs according to the grid $\{1,2,3,4,5,10,20,50,100,200,300,400\}$, and run the stochastic algorithms using the same setup described in the previous subsection. In all subfigures, the x-axis represents the number of epochs, while the y-axis is the AUC classification accuracy on the test data.  

The results show that the proposed NSAUC converges to the optimal solution on all datasets. We can also see that the AUC performance of NSAUC outperforms its non-averaging variant NSVMSGD2 on four datasets (i.e., ijcnn1, cod-rna, acoustic, and connect-4), while its training time is on par with that of NSVMSGD2. This indicates the effectiveness of incorporating the scheduled averaging technique. Furthermore, the AUC performance of NSAUC does not fluctuate with varying the number of epochs on all datasets. This implies that choosing the best number of epochs would be easy.    

In addition, we can observe that the AUC performance of NSOLAM does not show significant improvement after the first epoch. The reason is that NSOLAM reaches a local minimum (i.e., a saddle point) in a single pass and gets stuck there.  

\section{Conclusion and Future Work}
In this paper, we have proposed scalable batch and stochastic nonlinear AUC maximization algorithms. The proposed algorithms optimize linear classifiers on a finite-dimensional feature space constructed via the k-means Nystr\"{o}m approximation. We solve the proposed batch AUC maximization algorithm using truncated Newton optimization, which minimizes the pairwise squared hinge loss function. The proposed stochastic AUC maximization algorithm is solved using a first-order gradient descent that implements scheduled regularization update and scheduled averaging to accelerate the convergence of the classifier. We show via experiments on several benchmark datasets that the proposed AUC maximization algorithms are more efficient than the nonlinear kernel AUC machines, while their AUC performances are comparable or even better than the nonlinear kernel AUC machines. Moreover, we show experimentally that the proposed stochastic AUC maximization algorithm outperforms the state-of-the-art online AUC maximization methods in terms of AUC classification accuracy with a marginal increase in the training time for some datasets. We demonstrate empirically that the proposed stochastic AUC algorithm converges to the optimal solution in a few epochs, while other online AUC maximization algorithms are susceptible to suboptimal convergence. In the future, we plan to use the proposed algorithms in solving large-scale multiple-instance learning.

\bibliographystyle{splncs04.bst}

\end{document}